\documentclass[letterpaper, 10 pt, conference]{ieeeconf}  

\IEEEoverridecommandlockouts                              

\usepackage{mathtools}
\usepackage{amsmath}
\usepackage{pdfpages}
\usepackage{cite}
\usepackage{hyperref}
\usepackage[normalem]{ulem}

\usepackage{xcolor}

\usepackage[ruled,vlined]{algorithm2e}

\DeclareMathOperator*{\argmax}{argmax} 

\title{\LARGE \bf
Bridging Visual Perception with Contextual Semantics for Understanding Robot Manipulation Tasks
}

\author{Chen Jiang$^{\dagger}$ and Martin Jagersand$^{\dagger}$
\thanks{$^{\dagger}$Authors are with Department of Computing Science,
        University of Alberta, Edmonton AB., Canada, T6G 2E8.
        { 
           \tt\small \{cjiang2, mj7\}@ualberta.ca
        }
        }%
}

\begin{document}

\maketitle
\thispagestyle{empty}
\pagestyle{empty}

\begin{abstract}

Understanding manipulation scenarios allows intelligent robots to plan for appropriate actions to complete a manipulation task successfully. It is essential for intelligent robots to semantically interpret manipulation knowledge by describing entities, relations and attributes in a structural manner. In this paper, we propose an implementing framework to generate high-level conceptual dynamic knowledge graphs from video clips. A combination of a Vision-Language model and an ontology system, in correspondence with visual perception and contextual semantics, is used to represent robot manipulation knowledge with Entity-Relation-Entity (E-R-E) and Entity-Attribute-Value (E-A-V) tuples. The proposed method is flexible and well-versed. Using the framework, we present a case study where robot performs manipulation actions in a kitchen environment, bridging visual perception with contextual semantics using the generated dynamic knowledge graphs.

\end{abstract}

\section{INTRODUCTION}
Traditionally, robots are controlled using motion paths pre-defined by coordinate-based programming or kinestatic teach-in demonstration. Modules involving SLAM, visual servoing and deep learning have increased the autonomous capability of intelligent robotics. As a result, robots nowadays are demonstrating much smarter behaviors through their capabilities to semantically process information on demand. Still, it is an open problem
to introduce automation for robots in daily manipulation tasks, which involves decomposing and analyzing various semantic concepts like a human can.

A variety of approaches towards using task semantic information for intelligent robotics have been studied, for example, from processing semantic information in visual scene \cite{yang2015learning, aditya2018image, nguyen2018translating}, to structuring taxonomy of the robot manipulation actions and activities \cite{bessler2018owl, paulius2019manipulation, mokhtari2019learning}. It is necessary that, an intelligent robot needs to systematically enact smart behaviors conditioned on the semantic concepts presented at the manipulation environment in real time. As such, an expressive and instructive mechanism is required to decompose and analyze the contextual semantics in activities. A unified modeling of vision, language and knowledge is needed over the fact that information is connected, convertible and accessible. 

In this paper, we propose an implementing framework to process robot manipulation concepts by bridging visual perception with contextual semantics. The intelligent robot first summarizes the presented semantic concepts by "watching" one single video clip from the manipulation scene. Given the semantic concepts from visual perception, the robot 
consults static, pre-programmed commonsense knowledge using an ontology system, and
completes the semantic concepts perceived into a dynamic knowledge graph, describing the properties of the semantic concepts of the video in detail. Our main contributions are:

\begin{itemize}
\item We present an implementing framework to generate a high-conceptual knowledge graph by taking a video clip from the dynamic manipulation scene. The generated dynamic knowledge graph summarizes the semantic concepts of the video clip in a taxonomic manner, under a life-long robot knowledge domain. 

\item We discuss key concepts and methods of our framework. Visual perception dynamically infers the scene of manipulation into a skeleton Entity-Relation-Entity (E-R-E) knowledge tuple, while multiple sets of Entity-Attribute-Value (E-A-V) knowledge tuples complete the skeleton into a full dynamic knowledge graph.

\item We present a case study where on-scene dynamic concepts of manipulation for robot need to be constantly inferred. The advantages and drawbacks of our implementing framework are then discussed.

\end{itemize}

The rest of the paper is organized as follows: section II summarizes the recent advances in intelligent robotics with vision-language and semantic knowledge; section III presents the implementing framework along with key definitions. Enabling methods for our framework are discussed in section IV; and we present our case study in section V; final conclusion and potential future work are summarized in section VI. 

\section{RELATED WORK}

\subsection{Vision and Language in Intelligent Robot}
There has been multiple studies combining visual perception with semantic language in intelligent robotics. Early studies propose to combine object detection with semantic language parser, generating grammar trees to describe manipulation actions \cite{yang2014manipulation, yang2015learning, yang2015robot, zhang2019learning}. Recent studies have adapted deep learning based methods, where CNN-RNN based architectures are used to caption manipulation actions into language sentences for a variety of robotic applications \cite{nguyen2018translating, nguyen2019v2cnet, yang2018learning, jiang2020understanding}. Other studies include the utilization of semantic scene graphs, representing an overall understanding over visual perception \cite{aditya2018image, armeni20193d}. Still, for intelligent robot to conduct action planning, capturing semantic actions alone is not enough. It is mandatory to capture the change of concepts themselves, which are the end results of executing those semantic actions. We formulate this factor into our implementing framework specifically.

\subsection{Semantic Knowledge in Intelligent Robot}
Multiple studies have considered ways to describe semantic knowledge in order to regulate robotic behaviors. One way is to construct taxonomic structures like ontology, constraining a variety of robotic behaviors into a specific knowledge domain \cite{lim2010ontology, bessler2018owl, nyga2018grounding, paulius2019manipulation, diab2019ontology, jiang2020understanding}. Other studies use representations like Finite State Machines (FSMs) \cite{de2019cognitive}, semantic graphs \cite{ramirez2014automatic, fox2019multi, mokhtari2019learning} or behavior trees \cite{marzinotto2014towards} to describe and structure robotic actions and activities. While those semantic structures are versatile in robotic action planning, most of them remain static and do not consider the dynamic changes presented on scene. In our framework, we combine visual perception with contextual semantics to dynamically process on-scene semantic knowledge.

\section{Implementing Framework Proposed}
Given a visual observation of the manipulation scene, we aim to describe the manipulation concepts and their internal relational structures into a dynamic knowledge graph. In this section, we first discuss how to combine visual perception with contextual semantics, generating from skeleton to full form of the dynamic knowledge graph. We then propose our implementing framework, enabling semantic understanding for intelligent robot.

\subsection{Visual Perception}
The fundamental role of visual perception is to summarize the entity and relation knowledge in the video content, serving as the skeleton for the dynamic knowledge graph. To describe concepts and their associated relations, we use Entity-Relation-Entity (E-R-E) tuples. The E-R-E tuple explicitly describes things and their probable action relationships. Entities are denoted as classes of objects or concepts in a knowledge domain, such as \textit{Centric}, \textit{Mug}, \textit{RobotArm}, etc. Relations are denoted as the semantic relations that connect any two entities, such as \textit{to}, \textit{on}, \textit{hold}, \textit{pour}, etc. An E-R-E tuple connects at least two entities with at least one presenting semantic relation in the manipulation scene exactly, such as static relations like \textit{RobotArm hold CentricMug} and \textit{CentricMug on Desk}, or action relations like \textit{RobotArm grasp PlasticBottle} and \textit{RobotArm pour ColdWater}, etc.

\subsection{Contextual Semantics}
Given the skeleton of the dynamic knowledge graph as one or more E-R-E tuples, we aim to capture the characteristics of the presenting entities in detail. We use Entity-Attribute-Value (E-A-V) tuples to describe the property of the associated entity with user-specified value. An E-A-V tuple can represent intrinsic property, for example, a mug's "material", "color", etc, and extrinsic property, for example, if a mug is "graspable" by a "named robot manipulator", if a mug can contain "hot water", etc. E-A-V tuples allow us to capture and distinguish the properties of any entity and to assert certain restrictions given different attribute values. By merging E-R-E tuple with a set of E-A-V tuples, the skeleton of the dynamic knowledge graph is fully completed.

\subsection{Workflow of the Framework}
\begin{figure}[h!]
\centering
    \includegraphics[scale=0.27]{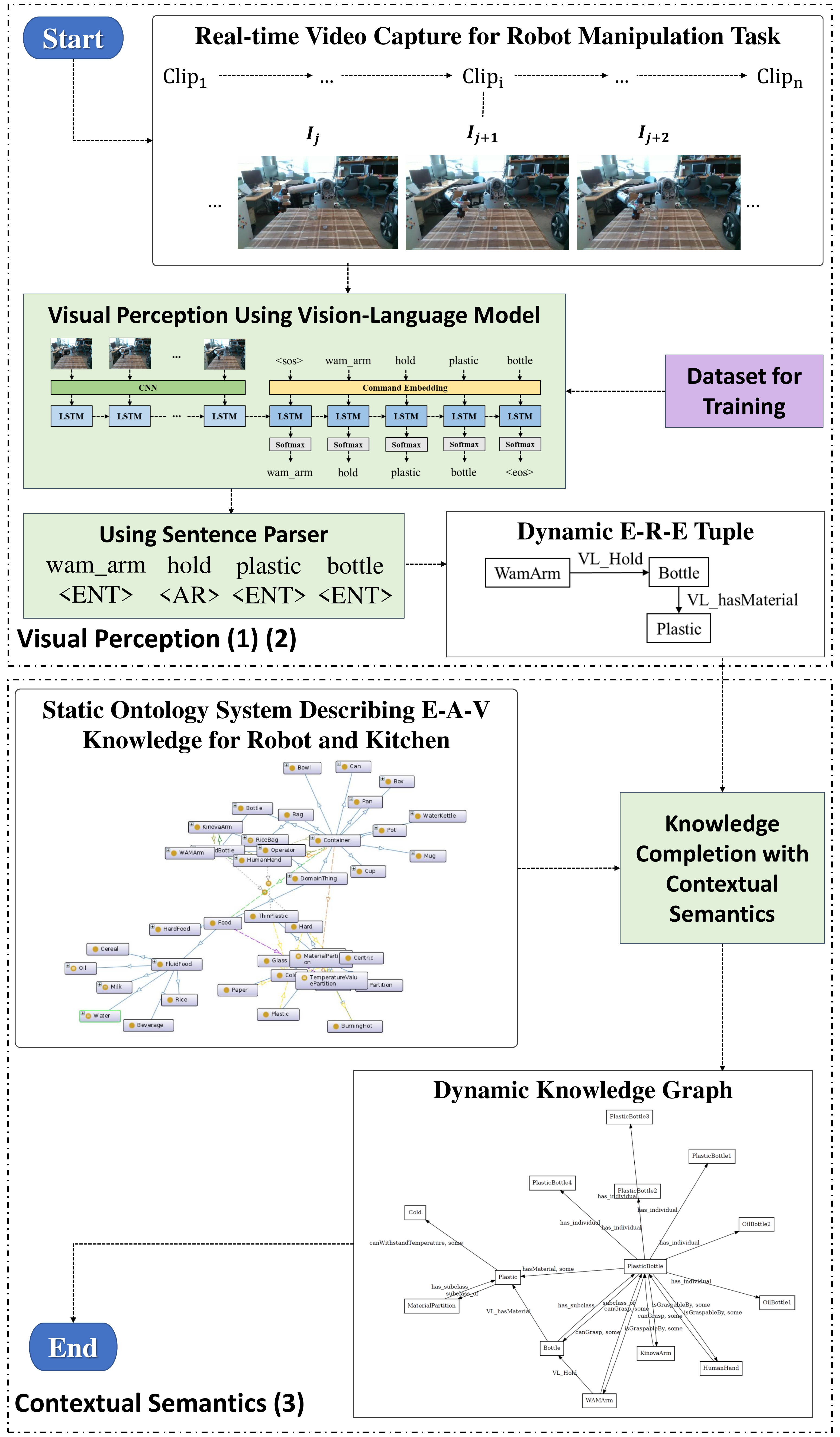}
    \caption{Overview of our implementing framework for robot semantic understanding procedure. }
     \label{fig:framework}
\end{figure}
Figure \ref{fig:framework} presents the implementing framework for robot semantic understanding procedure. The logic configuration of the framework can be highlighted as follows: 

\begin{enumerate}
\item \textbf{Robotic Vision Settings:} A live static camera is setup to frontally observe the scene where manipulation actions can possibly take place. A manipulator can impose certain actions on some objects of interest to complete a task of manipulation. The process of observation can be segmented into multiple video clips. Entities and relations between them can be inferred as E-R-E semantic sentences for each of the video clip.

\item \textbf{E-R-E Knowledge Generation:} A pre-trained Vision-Language model takes a segmented video clip from robotic vision as input, captioning one or more semantic sentences. A sentence parser is then followed to process the generated semantic sentences into E-R-E tuples, serving as the skeleton of the dynamic knowledge graph. 

\item \textbf{Dynamic Knowledge Graph Completion:} By observing all entities and their associated relations in a dataset of robot manipulation task videos, an ontology system can be constructed to store those manipulation knowledge as machine-interpretable definitions of concepts. Given a keyword of entity, the ontology system can generate multiple E-A-V tuples associated with the querying entity. With those E-A-V tuples, E-R-E skeleton can be completed into a full dynamic knowledge graph.

\end{enumerate}

In order to successfully generate the dynamic knowledge graph describing robot manipulation knowledge, some key enabling methods need to be developed, including (1) Vision-Language model for dynamic E-R-E knowledge inference; (2) ontology system for E-A-V knowledge inference; and (3) algorithm for the final dynamic knowledge graph completion. We discuss each of the methods in the next section.

\section{Key Enabling Methods}
\subsection{Vision-Language Model}
\begin{figure}[h!]
\centering
    \includegraphics[scale=0.35]{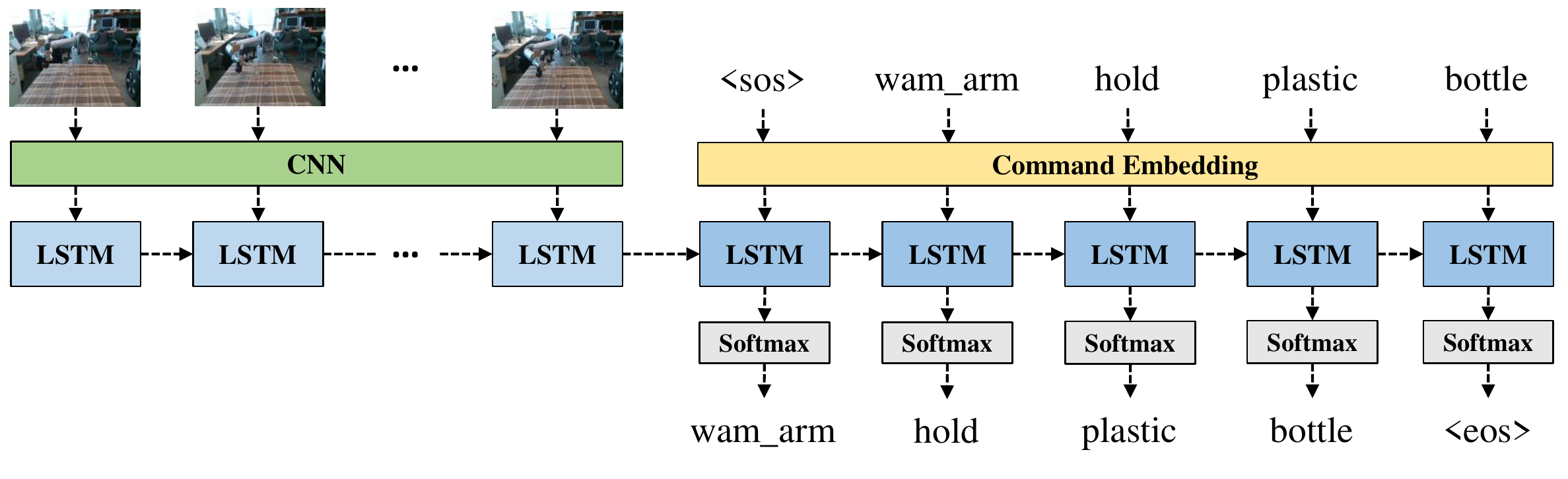}
    \caption{seq2seq architecture used to caption semantic sentence given one video clip. }
     \label{fig:seq2seq}
\end{figure}
To infer for the E-R-E knowledge presented in the duration of a video clip, an End-to-End Vision-Language model can be trained. The Vision-Language model used is dependent on the seq2seq architecture \cite{Sutskever2014SequenceTS} to process sequential data, presented in Figure \ref{fig:seq2seq}. Given a video clip consisting of $n$ image frames $C = (I_1, I_2, ..., I_n)$, a convolutional neural network (CNN) pretrained on ImageNet is first employed to extract image features as $Z = (z_1, z_2, ..., z_n)$. A Long Short-Term Memory network (LSTM) is then used to encode the visual features into an encoding vector representation $v$. One time step of LSTM update is shown below, given the visual feature $z_{t}$ at timestamp $t$, the hidden state $h_{t-1}$ and the memory cell state $c_{t-1}$ at previous timestamp $t-1$:

\begin{align*}
i_t &= \sigma (W_{ii} z_t + b_{ii} + W_{hi} h_{t-1} + b_{hi}) \\ 
f_t &= \sigma (W_{if} z_t + b_{if} + W_{hf} h_{t-1} + b_{hf}) \\
\tag{1}
g_t &= \tanh (W_{ig} z_t + b_{ig} + W_{hg} h_{t-1} + b_{hg}) \\ 
o_t &= \sigma (W_{io} z_t + b_{io} + W_{ho} h_{t-1} + b_{ho}) \\ 
c_t &= f_t * c_{t-1} + i_t * g_t \\ 
h_t &= o_t * \tanh (c_t)
\end{align*}

\noindent where $\sigma$ is the sigmoid function, and $i_t$, $f_t$ and $o_t$ represent the input state, forget state and output state at timestamp $t$. $v = (h_{t_n}, c_{t_n})$ is the last hidden state and the memory cell state of the encoding LSTM. Another LSTM is then employed to decode an E-R-E semantic sentence sequentially, conditioned on the encoding vector $v$: 
$$
p(s_1 ..., s_K|z_{t_1}, ..., z_{t_n}) = \displaystyle\sum_{k=1}^{K} p(s_k|v, s_1, ..., s_{k-1}) \eqno{(2)}
$$
\noindent where $s_1 ..., s_K$ is the sequence of E-R-E semantic sentence with a maximum length of $K$. $p(s_k|v, s_1, ..., s_{k-1})$ is represented with a softmax over all the language tokens. The maximum log-likelihood objective to optimize the model parameters $\theta$ can be expressed as:
$$
\argmax_\theta \sum_{(Z,S)}\log p(S|Z; \theta) \eqno{(3)}
$$

\subsection{Ontology System}
\begin{figure}[h!]
\centering
    \includegraphics[scale=0.39]{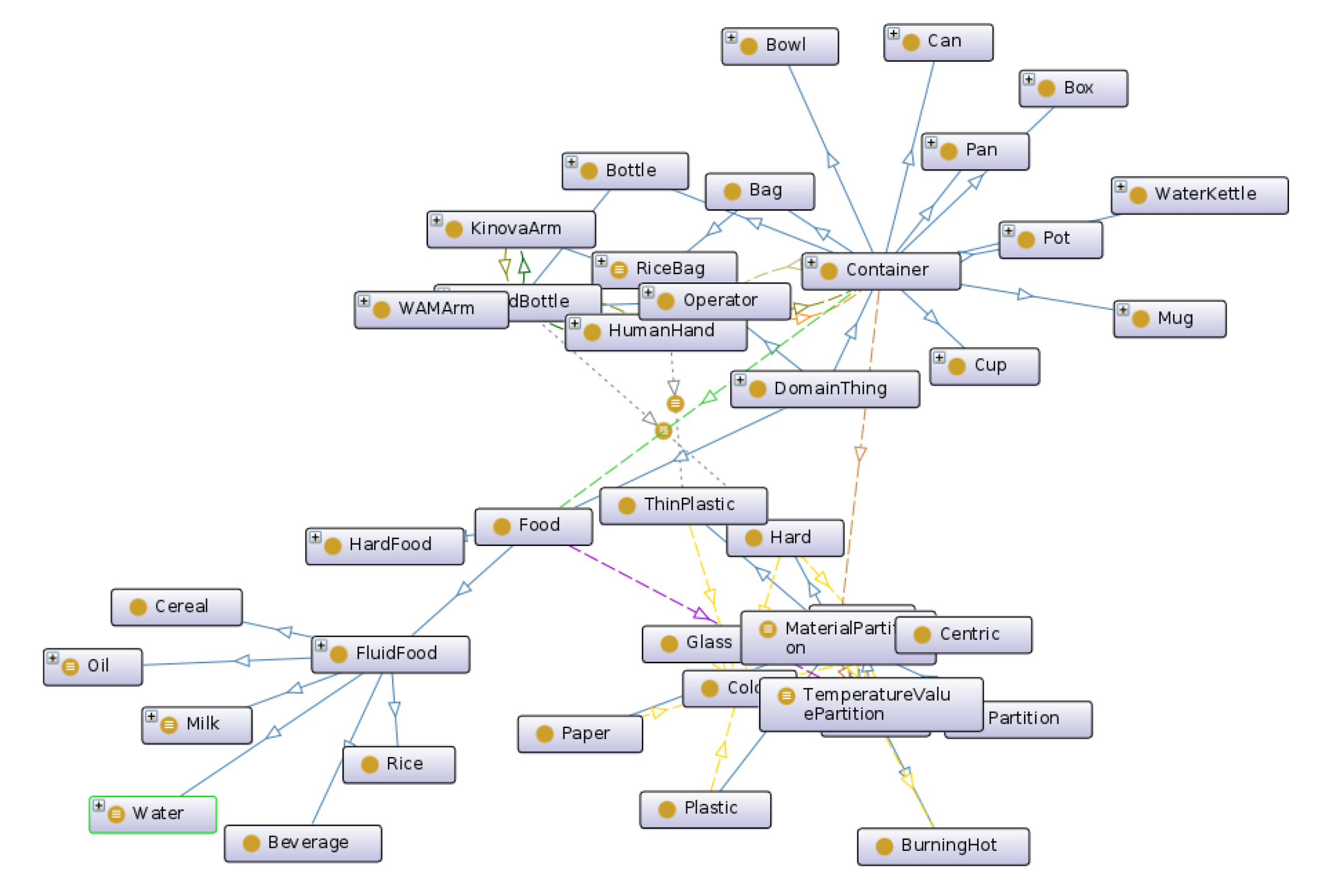}
    \caption{Visualization of the ontology system in Protege for robot manipulation knowledge.}
     \label{fig:ontology}
\end{figure}
The ontology system is constructed to store and process formal explicit description over the concepts of manipulation. An ontology tree arranges classes or concepts from robotic workspace in a taxonomic hierarchy. Intrinsic and extrinsic properties are then imposed and stored as object properties, defining relationships among classes. Figure \ref{fig:ontology} shows an example of the ontology tree for robot manipulation concepts visualized in Protégé \cite{musen2015protege}.

The constructed ontology system consists of 3 main classes: (a) "DomainThing", where kitchen objects in the manipulation contexts, like \textit{manipulator}, \textit{container}, \textit{liquid food}, etc, are presented; (b) "DomainPartition", where semantic proportional partitions like \textit{temperature}, \textit{emptiness}, etc, are presented. (c) "DomainTask", where configurations describing manipulation task scenarios and their corresponding action concepts are presented, including \textit{GraspingActionTask}, \textit{PouringActionTask}, corresponding to \textit{GraspingAction}, \textit{PouringAction}, etc. Intrinsic properties describe internal object attributes such as \textit{hasMaterial}, \textit{canWithstandTemperature}, etc, while extrinsic properties describes abstract relations such as \textit{hasActionPresented}, \textit{hasContainerPresented}, etc.

\subsection{Algorithm for Dynamic Knowledge Graph Completion}
Given any pre-trained Vision-Language model, an video clip is inputted into the pre-trained model, captioning E-R-E knowledge as one or more semantic sentences. A sentence parser is then followed to generate entity and relation tags for each word token in the sentences, allowing the parser to further process the generated sentences into E-R-E tuples. For each entity presented in the E-R-E tuples, the ontology system is queried, returning E-A-V tuples which are further merged with the E-R-E tuples. The final dynamic knowledge graph is completed when all the entities are queried in the set of generated E-R-E tuples and all E-A-V tuples are merged with the E-R-E tuples. A generic template pseudo-algorithm for generating the Dynamic Knowledge Graph over a single video clip is shown in Algorithm \ref{fig:alg_dkg}. In addition, the algorithm can be further adapted to progressively integrate time attribute \cite{jiang2020understanding}.
\begin{algorithm}[]
\SetAlgoLined
\textbf{Inputs:} A video clip $C$. A Vision-Language Model $Model$. A static ontology tree $onto$.\\
\KwResult{Dynamic knowledge graph $G_{C}$ over the video clip $C$}
 initialize an empty $G_{C}$\;
 $S_{C}$ $\gets$ Model($C$)\;
 MERGE($G_{C}$, $S_{C}$)\;
 \For{$entity$ in $S_{C}$}{
    $S_{entity}$ $\gets$ QUERY($onto$, $entity$)\;
    MERGE($G_{C}$, $S_{entity}$)
 }
 \caption{Generate a Dynamic Knowledge Graph}
 \label{fig:alg_dkg}
\end{algorithm}

\section{CASE STUDY}
\subsection{Case Description}
Following the definitions, implementing framework and key enabling methods for robot manipulation task understanding discussed above, we study a real-world case scenario where on-scene dynamic concepts of manipulation need to be constantly observed and inferred in the robotic vision. A kitchen table is presented in front of a Intel RealSense D435i RGB-D camera with random empty and full containers on the table. A Barrett WAM robot is guided to execute a sequence of motions to complete the a full manipulation task of pouring liquid from the full to empty vessel. 

\subsection{Case Usage with Visual Perception}
\begin{table}[h]
\caption{Quantitative Evaluation Results with Baselines on robotic data from Robot Semantics Dataset. }
\label{result_vl}
\begin{center}
\begin{tabular}{c|c c c c c}
\hline
\textbf{Name} & \textbf{B-4} & \textbf{M} & \textbf{R} & \textbf{C}  \\
\hline
seq2seq-ResNet50 & \textbf{0.486} & \textbf{0.453} & 0.754 & \textbf{3.609} \\
\hline
seq2seq-ResNet101 & 0.436 & 0.425 & 0.737 & 3.400 \\
\hline
seq2seq-ResNet152 & 0.420 & 0.435 & \textbf{0.755} & 3.542 \\
\hline
seq2seq-VGG16 & 0.383 & 0.431 & 0.732 & 3.322 \\
\hline
seq2seq-VGG19 & 0.369 & 0.404 & 0.720 & 3.229 \\
\hline
\end{tabular}
\end{center}
\end{table}

To enable visual perception by means of generating E-R-E tuples in real time, offline preparation is mandatory to acquire a pre-trained Vision-Language model. To do this, we use the robotic portion of the Robot Semantics dataset \cite{jiang2020understanding} to train and evaluate our Vision-Language model. For Robot Semantics dataset, E-R-E tuples are defined by action relations between the robot manipulator and some manipulation object of interest.

\subsubsection{Training} The Vision-Language model using the mentioned seq2seq architecture is trained in an end-to-end fashion. We evaluate different backbone CNN for frame feature extraction: ResNet50, ResNet101, ResNet152, VGG16 and VGG19. The word embedding vector and the weights for LSTM are all randomly initialized, and teacher-student forcing is used, where the next ground truth word token of the E-R-E semantic sentence will always be used as the training target. Training is done with Adam for 100 epochs with a learning rate of 0.0001. The video stream sampling method proposed in Jiang et al \cite{jiang2020understanding} is employed here, sequentially generating 2453 clips for training. 

\subsubsection{Evaluating} The LSTM language generator is initialized into zero states. The generation process will not terminate until the language generator reaches the end token “eoc” prediction or the maximum sentence generating length, which we choose as 15. We validate the robustness of our trained Vision-Language model on the evaluation divisions of the Robot Semantics dataset, generating 2103 clips for evaluation. Standard machine translation and language generation metrics, including BLEU 1-4, METEOR, ROUGE-L, and CIDEr, are reported in table \ref{result_vl}. All scores are computed with the coco-evaluation code \cite{chen2015microsoft}.

\subsubsection{Using} Given a video clip of the robot ”pouring" manipulation scene as an input, our pre-trained Vision-Language model can successfully capture the semantic concepts presented on scene, and summarizes them into an interpretable E-R-E tuple. The E-R-E tuple serves as the skeleton of the dynamic knowledge graph. 

\subsection{Case Usage with Contextual Semantics}
\begin{figure*}[!tp]
\centering
    \includegraphics[scale=0.45]{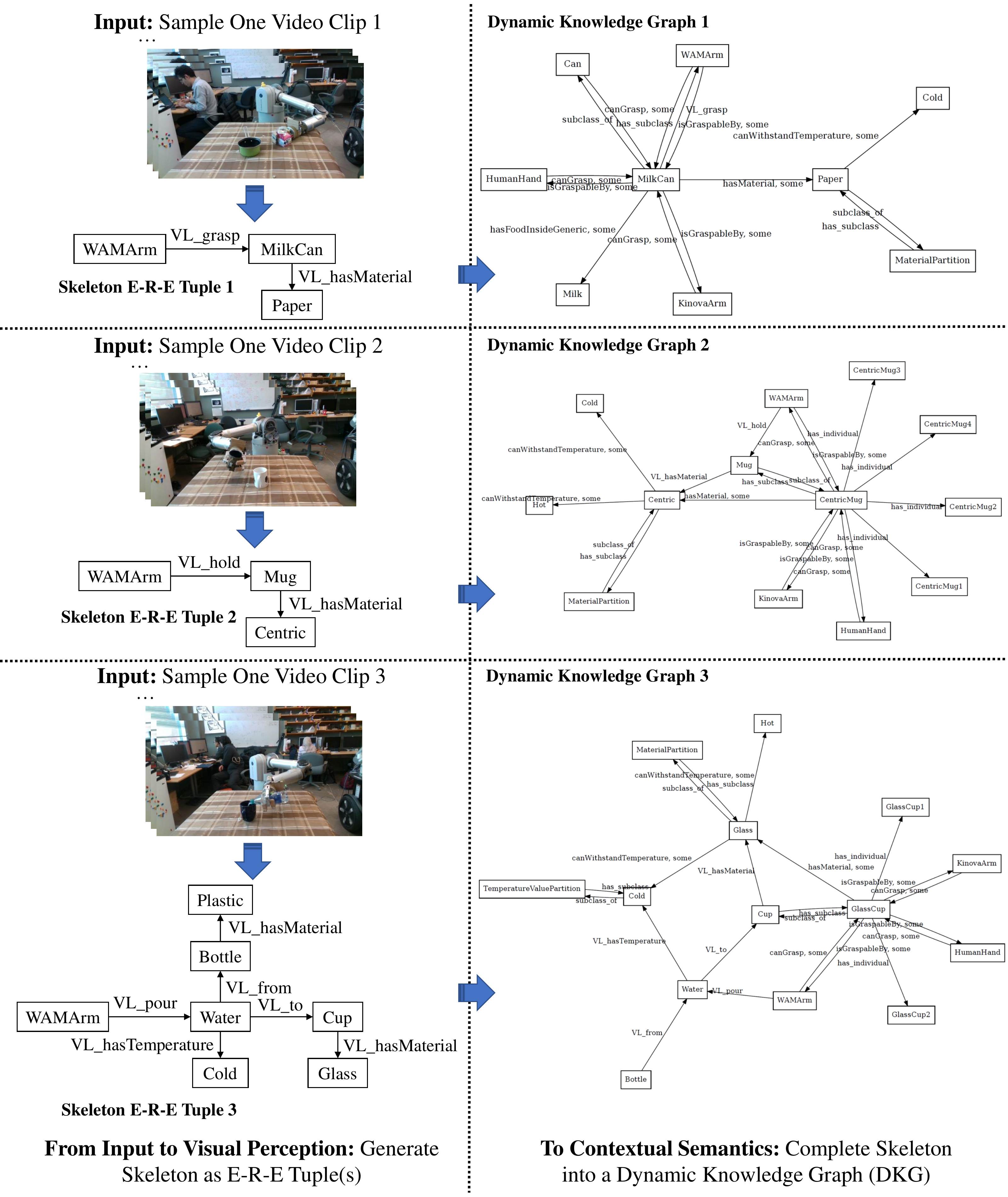}
    \caption{Examples of some generated dynamic knowledge graphs following our implementing framework. Three independent video clips from three scenarios are shown.}
     \label{fig:result_dkg}
\end{figure*}

The specialized ontology system is constructed in association with the Robot Semantics dataset using Protégé \cite{musen2015protege} and Owlready2 \cite{lamy2017owlready}. The above E-R-E tuple generated through visual perception represents a skeleton of the dynamic knowledge graph. For each entity inside the E-R-E tuple, the ontology system successfully generates relevant E-A-V tuples associated to the queried entity. E-A-V tuples are further assembled with skeleton E-R-E tuples into the full dynamic knowledge graph with generic Algorithm \ref{result_vl}. Some examples of our dynamically generated knowledge graphs are shown in Figure \ref{fig:result_dkg}. 

\subsection{Discussion}
While the generated dynamic knowledge graph shows promising details in regards of inputted robot manipulation tasks, there are still a few facts to be discussed. The details of the dynamic knowledge graph are extremely susceptible to the quality of the E-R-E knowledge being produced by the visual perception, as E-R-E tuples directly influence the entities to be queried. As such, it is essential to support the implementing framework with a high quality pre-trained Vision-Language model. And consequentially, the quality of the dynamic knowledge graph suffers if the Vision-Language model performs poorly with respect to the manipulation scene, or if the extracted CNN features are not robust enough. The quality of the E-A-V tuples are constrained by a static knowledge domain. This can be beneficial when the world knowledge of robot manipulation task is focused and consistent. However, things can become tricky to maintain when a great deal of static commonsense knowledge is added into the static domain.

Never the less, the implementing framework achieves the desirable outcomes in our case scenario. And it is straightforward to apply this framework into further robotic scenarios. For example, the dynamic knowledge graph can be extremely helpful to interpret the human intention for robot to assist the human in daily manipulation tasks. Given the specific scene information and human-robot configuration, for instance, an empty mug and a bottle with water, the robot will be able to semantically understand the concept of emptiness with contextual semantics. Therefore, robot can autonomously intend for its manipulation task execution of pouring water from the full bottle into an empty mug, and handing in the mug to the disabled. 

\section{CONCLUSIONS}
In this paper, we describe an implementing framework and its correspondent key enabling methods to represent semantic knowledge in robot manipulation task by combining visual perception with contextual semantic. Dynamic knowledge graphs concerning the semantic context of the robot manipulation task are generated from skeleton to full using the pre-trained Vision-Language model and the ontology system. The preliminary case study and analysis show promising results over knowledge representation for intelligent robots. The proposed methodology is compatible and adaptive in numerous applying scene requiring robot semantic understanding. It is also clear that some further strategies, such as, visual attention mechanism, intention based action planning, deep fusion of neural computing and symbolic reasoning, etc, can power the implementing framework proposed. The idea of utilizing knowledge of manipulation with intention is critical in intelligent Human-Robot interaction especially, and we plan to have further investigation.






{
\bibliographystyle{IEEEtranS}
\bibliography{citation}
}

\end{document}